\documentclass[final]{cvpr}

\usepackage{times}
\usepackage{epsfig}
\usepackage{graphicx}
\usepackage{amsmath}
\usepackage{amssymb}

\usepackage{mathrsfs}
\usepackage{bm}
\usepackage{algorithm, algorithmic}
\usepackage{mathdots}
\usepackage{multirow}
\usepackage[american]{babel}
\usepackage{microtype} 
\usepackage{array}
\usepackage{setspace}
\usepackage{threeparttable}

\usepackage[pagebackref=true,breaklinks=true,colorlinks,bookmarks=false]{hyperref}

\def\cvprPaperID{16} 
\def\confYear{CVPR 2021}

\begin{document}

\title{Improving Adversarial Transferability with Gradient Refining}


\author{First Author\\
Institution1\\
Institution1 address\\
{\tt\small firstauthor@i1.org}
\and
Second Author\\
Institution2\\
First line of institution2 address\\
{\tt\small secondauthor@i2.org}
}

\author{
Guoqiu Wang\footnotemark[1], Huanqian Yan\footnotemark[1], Ying Guo\footnotemark[1], Xingxing Wei\footnotemark[2] \\
Beijing Key Laboratory of Digital Media, Beihang University, Beijing, China\\
{\tt\small  \{wangguoqiu, yanhq, yingguo, xxwei\}@buaa.edu.cn}
}

\maketitle
 
\renewcommand{\thefootnote}{\fnsymbol{footnote}}
\footnotetext[1]{Equal contribution.} 
\footnotetext[2]{Corresponding author.}

\begin{abstract}
   Deep neural networks are vulnerable to adversarial examples, which are crafted by adding imperceptible perturbations to original images. Most existing adversarial attack methods achieve nearly 100\% attack success rates under the white-box setting, but only achieve relatively low attack success rates under the black-box setting. To improve the transferability of adversarial examples for the black-box setting, several methods have been proposed, e.g., input diversity, translation-invariant attack, and momentum-based attack. In this paper, we propose a method named Gradient Refining, which can further improve the adversarial transferability by correcting negative gradients introduced by input diversity through multiple transformations. Our method is generally applicable to many gradient-based attack methods combined with input diversity. Extensive experiments are conducted on the ImageNet dataset and our method can achieve an average transfer success rate of 82.07\% for three different models under single-model setting, which outperforms the other state-of-the-art methods by a large margin of 6.0\% averagely. And we have applied the proposed method to the competition \textit{CVPR 2021 Unrestricted Adversarial Attacks on ImageNet} and won the second place in attack success rates
   among 1558 teams.
\end{abstract}

\section{Introduction}

Though deep neural networks (DNNs) have achieved state-of-the-art performance on various vision tasks, including image classification \cite{krizhevsky2012imagenet,simonyan2014very}, object detection \cite{girshick2015fast,ren2015faster,redmon2016you} and semantic segmentation \cite{chen2017deeplab,long2015fully}, they are vulnerable to adversarial examples \cite{goodfellow2014explaining,szegedy2013intriguing} which are crafted by adding imperceptible perturbations to original images, making models output wrong predictions expected by attackers. 
The existence of adversarial examples has raised concerns in security-sensitive applications, e.g., self-driving cars \cite{liu2019perceptual} and face recognition \cite{guo2021meaningful}. And two related research directions, one is trying to improve the attack ability of adversarial examples, the other is to improve the robustness of DNNs against the adversarial examples, promote each other and develop together.

\renewcommand{\thefootnote}{\arabic{footnote}}

\begin{figure}[t]
\centering\includegraphics[width=0.48 \textwidth]{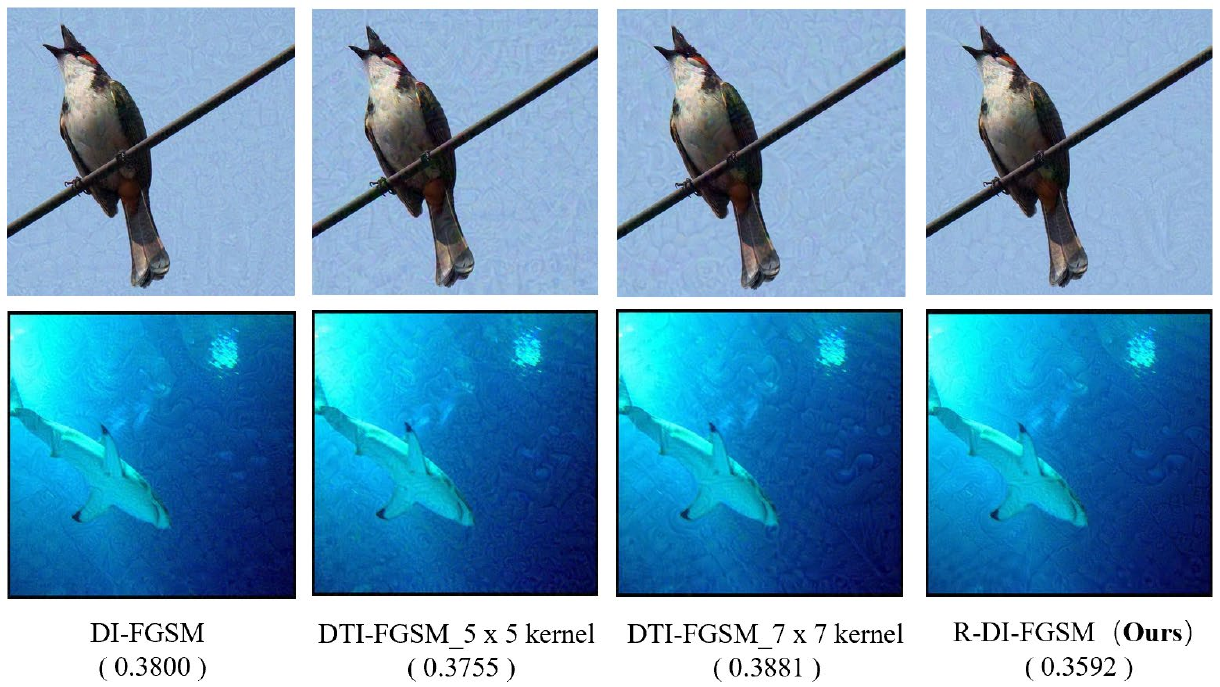}
\caption{\small Several adversarial examples of different attacks. The text denotes the corresponding attack method and LPIPS \protect\footnotemark[1] is shown in brackets. The transfer attack success rates are 67.10\%, 69.88\%, 69.92\% and 70.02\% respectively. The adversarial perturbation generated by our method in the fourth column shows better imperceptibility and transferability.}
\label{fig:ti}
\end{figure}
 \footnotetext[1]{LPIPS \cite{zhang2018unreasonable} is the perceptual similarity with original image (the lower the better).}

The generation of adversarial examples can serve as an important surrogate to evaluate the robustness of different models, and many methods, such as FGSM \cite{goodfellow2014explaining}, PGD \cite{madry2017towards}, and C\&W \cite{carlini2017towards}, have been proposed, which use the gradient information of the given model and are known as white-box attacks. Moreover, some works show that the adversarial examples have transferability \cite{liu2016delving, papernot2017practical}, which means the adversarial examples crafted for one model can attack a different model successfully. The existence of transferability makes black-box attacks,  where the attackers have no knowledge of the model structure and parameters, practical to the real-world applications and induces serious security issues.

Although above white-box attacks can achieve high success rates, they have low transferability. Recently, many methods have been proposed to improve the transferability, such as multi-model ensemble transferable attack \cite{liu2016delving}, input diversity \cite{xie2019improving}, translation-invariant attack \cite{dong2019evading}, and momentum-based attack \cite{dong2018boosting}. By attacking an ensemble of models, an example remains adversarial for multiple models, and is more likely to transfer to other black-box models. 
Input diversity applies random and differentiable transformations to the input images to create hard and diverse input patterns.
Translation-invariant attack smooths the gradient with a pre-defined kernel, making
adversarial examples less sensitive to the discriminative region of the white-box model being attacked. Momentum-based iterative attack integrates the momentum term into the iterative process, which can stabilize the update directions and escape from poor local maxima during the iterations, leading to more transferable adversarial examples.

Fundamentally, the over-fitting of the source model is one of the key reasons for the low transferability of white-box attacks \cite{xie2019improving}. Input diversity \cite{xie2019improving} and translation-invariant attack \cite{dong2019evading} mentioned above decrease the over-fitting by random input transformations or smoothing the gradient with a kernel to improve the transferability. However, the randomness introduced by image transformations in input diversity does not always play a positive role in improving the transferability and sometimes brings the random \textbf{negative gradient information}, which has a negative effect in improving the transferability. 
We propose  a method to neutralize the negative gradient information, which considers multiple perturbations from multiple random transformations of the same image comprehensively, retains the positive gradient information, and counteracts the negative gradient information to refine the final gradient. Combining our method with existing methods, such as translation-invariant attack and momentum-based attack, the transferability can be further improved. Extensive experiments demonstrate the effectiveness of our method and show that the generated perturbations have better imperceptibility (see Figure \ref{fig:ti}). 

In brief, the main contributions of this paper can be summarized as follows:
\begin{itemize}

\item 
We propose an effective method to improve the adversarial transferability, which counteracts the negative gradients generated during the input diversity process by comprehensively considering the gradients of multiple transformed images.

\item Our method has good scalability. It can be well integrated with existing attack strategies (e.g., I-FGSM \cite{DBLP:journals/corr/KurakinGB16}, MI-FGSM \cite{dong2018boosting}, TI-FGSM \cite{dong2019evading}) to further improve the adversarial transferability.

\item Extensive experimental results show that the proposed method can get a higher transfer success rate of 82.07\% on average, which outperforms the other mainstream methods by a large margin of 6.0\%. By applying the proposed method to the competition CVPR 2021 Unrestricted Adversarial Attacks on ImageNet, we ranked fourth overall, and won the second place in attack success rates among 1558 teams.



\end{itemize}


\section{Related Work}

In this section, we provide the background knowledge of adversarial attack as well as several classic methods which are designed to improve the attack ability and transferability of adversarial examples.

Given a classification network $f_{\theta}$ parameterized by $\theta$, let $(x, y)$ denote the original image and its corresponding ground-truth label, the goal of adversarial attacks is to find an example $x_{adv}$ which is in the vicinity of $x$ but misclassified by the network. In most cases, we use the $L_{p}$ norm to limit the adversarial perturbations below a threshold $\epsilon$, where $p$ could be $0$, $2$, $\infty$. This can be expressed as:
\begin{equation}\label{eq:untarget}
f_{\theta }(x_{adv})\neq y, \ s.t. \ \left \| x_{adv}-x \right \|_{p}\leq \epsilon
\end{equation}

\textbf{Input diversity iterative (DI) attack \cite{xie2019improving}} applies random transformations to the input images at each iteration to create hard and diverse input patterns, which brings randomness to the adversarial perturbations and improves the generalization of adversarial examples efficiently. This method can be formalized as:
\begin{equation}\label{eq:untarget}
x_{t+1}^{adv} = x_{t}^{adv} + \alpha \cdot sign(\bigtriangledown _{x}L(T(x_{t}^{adv}, p), y))
\end{equation}
where $\bigtriangledown _{x}L(T(x_{t}^{adv}, p), y)$ denotes the gradient of the loss function $L(\cdot, \cdot)$ w.r.t. the transformed image of adversarial example $x_{t}^{adv}$ generated in step $t$. $sign(\cdot)$ is the sign function and $\alpha$ denotes the step size. $T(\cdot, p)$ denotes the stochastic transformation function with probability $p$. If $p=1.0$, then only transformed inputs are used for the attack. Since the original inputs are not seen by the attackers, the generated adversarial examples tend to have much higher attack success rates on black-box models but lower success rates on white-box models.

\textbf{Translation-invariant iterative (TI) attack \cite{dong2019evading}} improves attack ability by smoothing the gradient of the image with a pre-defined kernel, making the adversarial example less sensitive to the discriminative region of the white-box model being attacked. This method can be formulated as:
\begin{equation}\label{eq:untarget}
x_{t+1}^{adv} = x_{t}^{adv} + \alpha \cdot sign(\bm{W} \ast  \bigtriangledown _{x}L(x_{t}^{adv}, y))
\end{equation}
where $\bm{W}$ denotes the pre-defined kernel. The size of $\bm{W}$ plays a key role for improving the transferability, and with the size increasing, the adversarial perturbations will be more perceptible (see Figure \ref{fig:ti}).

\begin{figure}[t]
\centering\includegraphics[width=0.475 \textwidth]{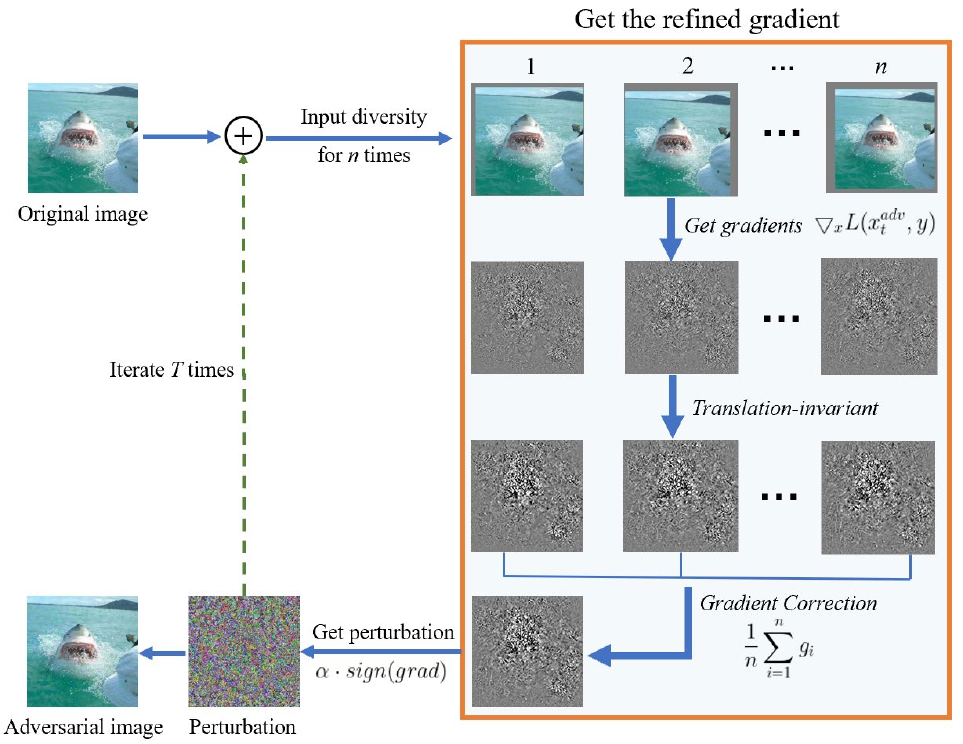}
\caption{\small The framework of R-DTI-FGSM algorithm.}
\label{fig:rng-dt}
\end{figure}

\textbf{Momentum-based iterative (MI) method \cite{dong2018boosting}} boosts adversarial attacks by integrating the momentum term into the iterative process to stabilize the update directions. The momentum-based iterative attack of FGSM (MI-FGSM) can be expressed as:
\vspace{-0.10cm}
\begin{equation}\label{eq:gt}
g_{t+1} = \mu \cdot g_{t} + \frac{\bigtriangledown _{x}L(x_{t}^{adv}, y)}{||\bigtriangledown _{x}L(x_{t}^{adv}, y)||_{1}}
\vspace{-0.15cm}
\end{equation}
\vspace{-0.15cm}
\begin{equation}\label{eq:xt+1}
x_{t+1}^{adv} = x_{t}^{adv} + \alpha \cdot sign(g_{t+1})
\end{equation}
where $g_{t}$ is the moment gradient and $\mu$ is the decay factor which is often set to $1.0$. MI-FGSM updates the moment gradient $g_{t+1}$ by Eq. (\ref{eq:gt}) and then updates $x_{t+1}^{adv}$ by Eq. (\ref{eq:xt+1}).

\vspace{-0.1cm} 
\section{Methodology}
\vspace{-0.06cm} 
Input diversity creates diverse input patterns and the image will be transformed once randomly with a certain probability during each iteration in the process of attacking an image, thus enhancing the transferability. 
But this inspires us a question: \textit {The randomness brought by input diversity does always positive? If not, how can we counteract negative randomness?}


Input diversity method brings random gradients and translation-invariant method magnifies the randomness. Though the random gradients reduce the over-fitting and make the adversarial examples more generalized and transferable, there exists random negative information of gradients, which inhibits the generalization and transferability of adversarial examples to some extent. 

\begin{algorithm}[t]
	\renewcommand{\algorithmicrequire}{\textbf{Input:}}
	\renewcommand{\algorithmicensure}{\textbf{Output:}}
	\caption{R-DTMI-FGSM Algorithm}
	\label{alg:1}
	\begin{algorithmic}[1]
	
		\REQUIRE {A classifier $f(\cdot)$ with loss function $L$, an original image $\bm x$ and ground-truth label $y$, correction number $n$, decay factor $\mu$, iterations $T$, the max perturbation $\epsilon$.}
		
		\ENSURE adversarial example $\bm x_{adv}$
		
		\STATE {\small $\alpha$ = $\epsilon$ / $T$};
		
		\STATE {\small $\bm g_{r}$ = $\bm 0$, $\bm x_{0}^{adv}$ = $\bm x$, $\bm g_{0} = \bm 0$};
		
		\FOR{$t$ = $0$  to $T\!-\!1$}
		    \STATE {\small Input $\bm x_{t}^{adv}$ to $f(\cdot)$ and obtain gradient $\bm {g_{r}}$ by

		    $ \bm {g_{r}} = \frac{1}{n}\sum_{i=1}^{n} (\bm{W} \ast \bigtriangledown _{x}L(T_{i}(x_{t}^{adv}, p), y))$ ;}

		    \STATE {\small Update $\bm g_{t+1}$ by $\bm g_{t+1}$ $ = \mu \cdot \bm g_{t} + \bm g_{r} $;}
		    \STATE {\small Update $\bm x_{t+1}^{adv}$ by
		    $x_{t+1}^{adv} = x_{t}^{adv} + \alpha \cdot sign(g_{t+1})$}
		    \STATE {\small $\bm g_{r}$ = $\bm 0$ ;}
		\ENDFOR
		\STATE {\small\textbf{return} $\bm x_{adv} = {x}_{T}^{adv}$ }
	\end{algorithmic}  
\end{algorithm}

To reduce the effects of random negative gradients introduced by input diversity, we design a method named Gradient Refining, which can enhance the transferability of adversarial examples efficiently. In the process of attacking an image, our method randomly transforms the image with several times during each iteration and calculates their respective gradients, then averages them to get a refined gradient, which preserves the gradients helpful for the transferability and counteracts the negative gradients for the attack. 

Gradient Refining can be combined with DI, DI-TI or DI-TI-MI, which called R-DI-FGSM, R-DTI-FGSM and R-DTMI-FGSM respectively. R-DTI-FGSM method can be formalized as:
\begin{equation}\label{eq:avg}
g_{r} = \frac{1}{n}\sum_{i=1}^{n} (\bm{W} \ast \bigtriangledown _{x}L(T_{i}(x_{t}^{adv}, p), y))
\end{equation}
\vspace{-0.25cm}

\vspace{-0.25cm}
\begin{equation}\label{eq:rng}
x_{t+1}^{adv} = x_{t}^{adv} + \alpha \cdot sign(g_{r})
\vspace{-0.10cm}
\end{equation}
where $n$ denotes the correction times of the gradient. With the increase of $n$, the attack ability will be improved, and when $n = 1$, it is DTI-FGSM. 

The framework of R-DTI-FGSM algorithm is in Figure \ref{fig:rng-dt}. For an original image, we use input diversity, which introduces randomness, for several times, to get multiple transformed images, and calculate their respective gradients. Translation-invariant method is used to amplify the randomness. Then we retain the positive information of gradients and counteract the negative gradients through average operation to get a refined gradient. We iterate the above process for $T$ times and get an adversarial image with high transferability.

For R-DTMI-FGSM algorithm, we use refined $g_{r}$ to update moment gradient $g_{t+1}$ and then update $\bm x_{t+1}^{adv}$ by Eq. (\ref{eq:xt+1}). The overall R-DTMI-FGSM algorithm is outlined in Algorithm \ref{alg:1}. In this paper, we show the algorithm and its experimental results under the constraint of  $L_{\infty}$ norm. The method can also be used in $L_{2}$ norm.



\begin{table*}[t]
\caption{The attack success rates (\%) under single-model setting and ensemble-model setting. $*$ indicates the white-box model being attacked. The right columns of $*$ for both setting show the transfer attack success rates in three different models and the average results for different attack methods. The improvement of average ASR by our method is shown in the brackets.}
\centering  
\resizebox{1 \textwidth}{14mm}{
\begin{tabular}{c|c|c|c|c|c|l|c|c|c|c|c|c}
\hline
\multicolumn{1}{l|}{}                 & \multicolumn{5}{c|}{Single-model attack}                                    & \textbf{} & \multicolumn{6}{c}{Ensemble-model attack}                                           \\ \cline{1-6} \cline{8-13} 
{Attack methods} & Res-101$^{*}$ & Den-161        & Inc-v4         & advIncRes-v2   & Average        & \textbf{} & Res-101$^{*}$ & VGG-16$^{*}$ & Den-161        & Inc-v4         & advIncRes-v2   & Average        \\ \cline{1-6} \cline{8-13} 
DI-FGSM                               & 99.96   & 72.58          & 62.28          & 37.28          & 57.38          & \textbf{} & 99.76   & 99.90  & 86.52          & 74.74          & 40.04          & 67.10          \\ \cline{1-6} \cline{8-13} 
R-DI-FGSM(\textbf {Ours})                       & 100.0   & 76.90          & 64.84          & 38.06          & 59.93 ($\uparrow$2.55)          & \textbf{} & 99.98   & 99.98  & 90.76          & 78.06          & 41.26          & 70.02 ($\uparrow$2.92)          \\ \cline{1-6} \cline{8-13} 
DTI-FGSM                              & 99.96   & 74.36          & 65.50          & 40.76          & 60.20          & \textbf{} & 99.82   & 99.92  & 88.48          & 77.60          & 43.58          & 69.88          \\ \cline{1-6} \cline{8-13} 
R-DTI-FGSM(\textbf {Ours})                      & 100.0   & 85.06          & 76.26          & 46.44          & 69.25 ($\uparrow$9.05)          & \textbf{} & 99.98   & 99.98  & 95.34          & 85.28          & 49.26          & 76.62 ($\uparrow$6.74)         \\ \cline{1-6} \cline{8-13} 
DTMI-FGSM                             & 99.90   & 90.50          & 82.64          & 55.30          & 76.14          & \textbf{} & 99.82   & 99.90  & 94.84          & 88.34          & 59.16          & 80.78          \\ \cline{1-6} \cline{8-13} 
R-DTMI-FGSM(\textbf {Ours})                     & 99.98   & \textbf{94.90} & \textbf{88.42} & \textbf{62.90} & \textbf{82.07} ($\uparrow$5.93) & \textbf{} & 99.92   & 99.96  & \textbf{97.00} & \textbf{91.16} & \textbf{64.82} & \textbf{84.32} ($\uparrow$3.54) \\ \hline
\end{tabular}} \label{tab:comparison111}
\end{table*}
\vspace{-0.1cm} 

\vspace{-0.1cm} 
\section{Experiments}

\begin{figure}[th]
\centering\includegraphics[width=0.47 \textwidth]{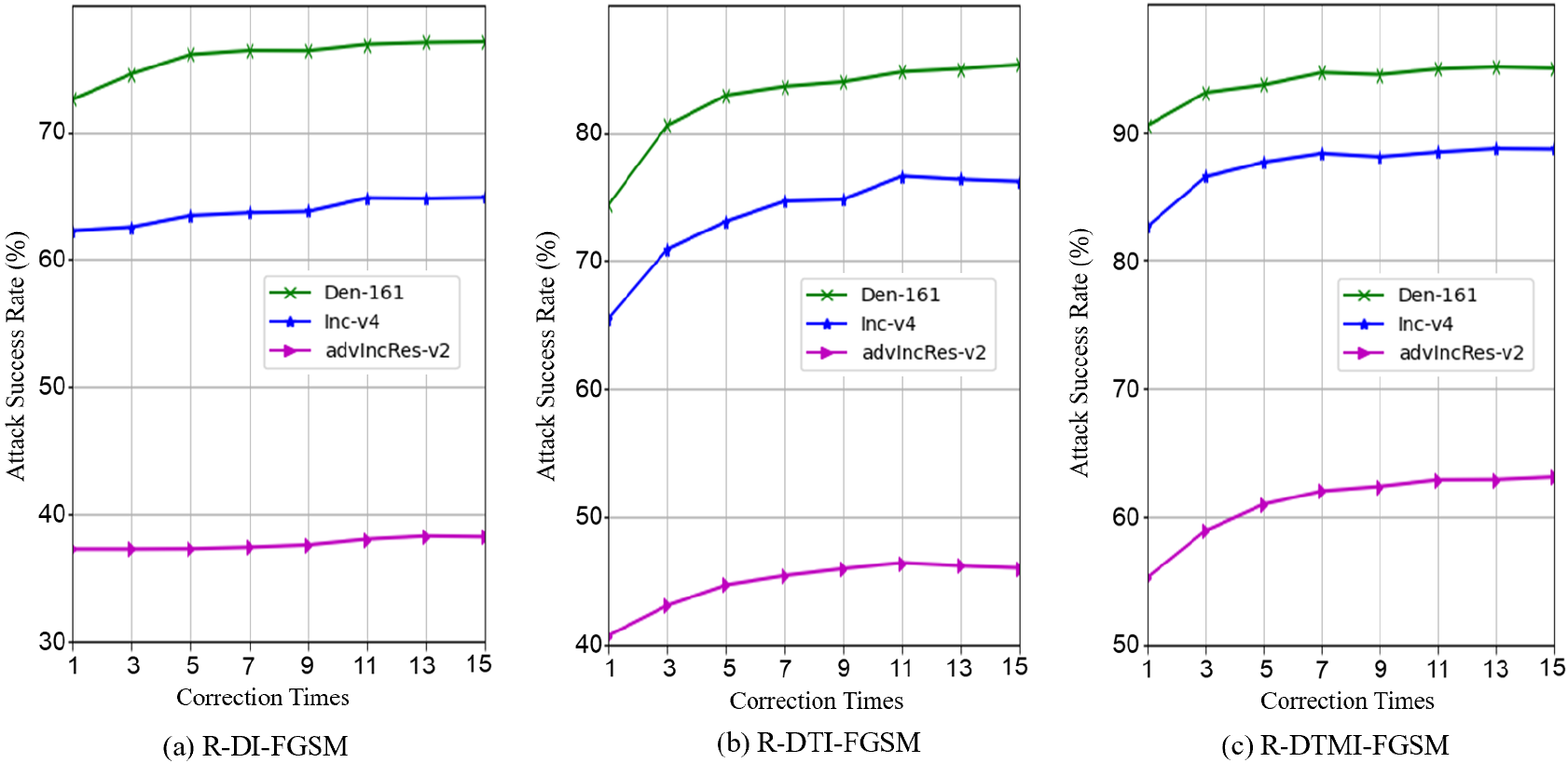}
\caption{\small The transfer attack success rates (\%) against Den-161, Inc-v4 and advIncRes-v2. The adversarial examples are generated on Res-101 model with the correction times ranging from 1 to 15.}
\label{fig:n}
\end{figure}

\subsection{Setup}

\textbf{Dataset:} We perform experiments on the dataset provided by alibaba group, which contains 5000 images selected from ImageNet \cite{deng2009imagenet}, five images in each class, and all these images are resized to 224 $\times$ 224.

\textbf{Networks:} We study four models with different structures and parameters, where three of them are normally trained models and the last one is the adversarially trained model. They are Resnet 101 (Res-101) \cite{he2016deep}, Denset 161 (Den-161) \cite{huang2017densely}, Inception v4 (Inc-v4) \cite{szegedy2017inception}, and adv-Inception Resnet v2 (advIncRes-v2) \cite{szegedy2017inception}.


\textbf{Implementation details:} We set the  maximum perturbation of
each pixel to be $\epsilon$ = 16 / 255 and the total iteration number $T$ is equal to $8$. For the stochastic transformation function $T(x,p)$, the probability $p$ is set to be $0.7$. 
The size of gaussian kernel $\bm{W}$ is $5 \times 5$.

\textbf{Metrics:} We use Attack Success Rate (ASR), which refers to the percentage of all images that can be misclassified by the target model, to evaluate different attack methods.

\vspace{-0.28cm} 
\subsection{Results}
\vspace{-0.1cm}
\subsubsection{The Effect of Correction Times}
\vspace{-0.22cm} 
The correction times $n$ plays a key role in improving the transfer attack success rates. If $n$ is equal to 1, our method will degenerate to their vanilla versions. Therefore, we conduct
an ablation study to examine the effect of correction times.
We attack the Res-101 model by R-DI-FGSM, R-DTI-FGSM, and R-DTMI-FGSM with different correction times, which range from 1 to 15 with a granularity 2, and the results are shown in Figure \ref{fig:n}. The transfer attack success rates continuously increase at first, and turn to remain stable after the correction times exceed 11. Considering attack ability and computational complexity, the correction times are set to 11 in the following experiments.

\vspace{-0.1mm}
\subsubsection{Attack Success Rates on Different Settings}
\vspace{-0.2mm}
We perform adversarial attacks using DI-FGSM, DTI-FGSM and DTMI-FGSM, and their extensions by combining with Gradient Refining attack method as R-DI-FGSM, R-DTI-FGSM and R-DTMI-FGSM. We report the attack success rates of the Gradient Refining based attacks with three baseline methods under single-model setting and ensemble-model setting. The results are shown in Table \ref{tab:comparison111}.

Under single-model attacks, the transfer attack success rates are improved by a large margin when using the proposed Gradient Refining on the listed attack methods. On average, the transfer attack success rates of Gradient Refining based attacks outperform the baseline attacks by 6.0\%. In particular, by using R-DTI-FGSM, the generated adversarial examples have 69.15\% average transfer attack success rates, which is improved by 9.05\% from the baseline method.

Under the ensemble-model setting, the results in table \ref{tab:comparison111} show that the transferability can also be improved by using Gradient Refining. Here we use two models, Res-101 \cite{he2016deep} and VGG-16 \cite{simonyan2014very}. R-DTI-FGSM improves the transfer attack success rates from the baseline method largely, because translation-invariant method magnifies the randomness brought by input diversity, and our method can correct the negative randomness and get a refined gradient to generate stronger adversarial perturbations. Besides, we visualize some adversarial images generated by DI-FGSM, DTI-FGSM and R-DI-FGSM in Figure \ref{fig:ti}. These visualization results show that the adversarial perturbations generated by our method have better imperceptibility.

\vspace{-0.2cm}  
\section{Conclusion}
\vspace{-0.1cm}  
In this paper, we proposed Gradient Refining method to generate adversarial perturbations that have higher transferability and better imperceptibility. By combining with input diversity, our method can be integrated into many existing attack methods. We conducted extensive experiments to validate the effectiveness. By using R-DTMI-FGSM algorithm, 
we can fool three different state-of-the-art models with an average success rate of 82.07\% under the single-model setting and higher success rates under the ensemble-model setting. Experimental results reveal the vulnerability of current models and inspire us to develop more robust models.

{\small
\bibliographystyle{ieee_fullname}
\bibliography{egbib}
}

\end{document}